\documentclass[conference]{IEEEtran}
\IEEEoverridecommandlockouts
\usepackage{cite}
\usepackage{amsmath,amssymb,amsfonts}
\usepackage{algorithmic}
\usepackage{graphicx}
\usepackage{textcomp}
\usepackage{xcolor}
\def\BibTeX{{\rm B\kern-.05em{\sc i\kern-.025em b}\kern-.08em
    T\kern-.1667em\lower.7ex\hbox{E}\kern-.125emX}}

\usepackage{times}
\usepackage{epsfig}
\usepackage{csquotes}
\usepackage{tabularx}

\usepackage{multirow}
\usepackage{multicol}

\usepackage{stfloats}

\usepackage[breaklinks=true,bookmarks=false]{hyperref}

\begin{document}

\title{
    An Investigation of Replay-based Approaches for Continual Learning
}

\author{
    \IEEEauthorblockN{1\textsuperscript{st} Benedikt Bagus}
    \IEEEauthorblockA{
        \textit{Fulda University of Applied Sciences} \\
        36037 Fulda, Germany \\
        benedikt.bagus@cs.hs-fulda.de
    }
    \and
    \IEEEauthorblockN{2\textsuperscript{nd} Alexander Gepperth}
    \IEEEauthorblockA{
        \textit{Fulda University of Applied Sciences} \\
        36037 Fulda, Germany \\
        alexander.gepperth@cs.hs-fulda.de
    }
}

\maketitle

\begin{abstract}
    Continual learning (CL) is a major challenge of machine learning (ML) and describes the ability to learn several tasks sequentially without catastrophic forgetting (CF).
    Recent works indicate that CL is a complex topic, even more so when real-world scenarios with multiple constraints are involved.
    Several solution classes have been proposed \cite{Lange2019}, of which so-called replay-based approaches seem very promising due to their simplicity and robustness.
    Such approaches store a subset of past samples in a dedicated memory for later processing: while this does not solve all problems, good results have been obtained.
    \\
    In this article, we empirically investigate replay-based approaches of continual learning and assess their potential for applications.
    Selected recent approaches as well as own proposals are compared on a common set of benchmarks, with a particular focus on assessing the performance of different sample selection strategies.
    We find that the impact of sample selection increases when a smaller number of samples is stored.
    Nevertheless, performance varies strongly between different replay approaches.
    Surprisingly, we find that the most naive rehearsal-based approaches that we propose here can outperform recent state-of-the-art methods.
\end{abstract}


\section{Introduction}\label{sec:int}
%

\subsection{Context}\label{sec:int:con}
In most application-oriented scenarios, data can only be processed as a stream, in small batches or sub-tasks, or is subject to concept drift over time.
For this reason, continual learning (CL) may be considered as a fundamental ability for modern machine learning (ML) models.
It may also lead to strong performance increases due to the huge amount of data that could be used for training models in a sequential fashion.
In contrast, deep neural networks (DNN) suffer from catastrophic forgetting (CF)\cite{McCloskey1989}, which is observed as a rapid drop in accuracy w.r.t. previous sub-tasks after a few iterations of training \cite{Pfuelb2019}.
Effectively, previously learned representations are irrevocably overwritten during the DNN parameter update process of the optimizer, unless this is explicitly mitigated.

\subsection{Broad strategies for mitigating CF}\label{sec:int:bro}
Among the proposed remedies to CF, three major directions may be distinguished \cite{Lange2019}: parameter isolation, regularization and replay.
The literature implementing these broad directions in specific ways is given and discussed in \ref{sec:int:rel}
\par\noindent\textbf{Parameter isolation} methods try to assign certain neurons or network resources to a single sub-task.
Dynamic architectures grow, e.g., with the number of tasks, so for each new sub-task new neurons are added to the network, which do not interfere with previously learned representations.
Fixed architectures must work with available neurons only and often restrict the number of used neurons per sub-task or reach a sparse representation through other concepts.
\par\noindent\textbf{Regularization} methods work by adding additional terms to the loss function.
Data-focused methods are mostly based on knowledge distillation and use old model outputs of new sub-task samples for regularizing further training steps.
Prior-focused methods typically define a notion of importance to a given sub-task (e.g., for weights) and then penalize changes to important parameters.
\par\noindent\textbf{Replay} methods store small subsets of previous sub-tasks, which can be used either for rehearsal (thus retaining training data of old sub-tasks) or as constraint generators.
If the retained samples are used for rehearsal, they are trained together with samples of the current sub-task.
If not, they are used directly to characterize valid gradients.
Instead of the dedicated memory, generative models can also be used.
In such cases, samples from previous sub-tasks are not stored but used to train a generator, e.g., generative adversarial networks (GANs) or variational autoencoders (VAEs).
These can subsequently produce an arbitrary number of samples from these sub-tasks.

\subsection{Related work}\label{sec:int:rel}
The problem of CF as the effect of sequential training is well known \cite{McCloskey1989} and was discussed quite early by \cite{Ring1998, Thrun1998}.
Currently, the field is expanding quickly, see \cite{Parisi2018, Hayes2018, Soltoggio2017, Lange2019} for reviews.
Systematic comparisons between different approaches to avoid CF are performed, see \cite{Kemker2017, Pfuelb2019}.
\\
Replay-based methods are a well known solution for CL.
However, multiple strategies have been proposed concerning sample selection and sample handling: sample selection decides which samples to retain in memory, and sample handling determines how to use them to avoid CL.
Typical representatives of replay methods are i.a. iCaRL \cite{Rebuffi2016}, GEM \cite{LopezPaz2017}, A-GEM \cite{Chaudhry2018}, GBSS \cite{Aljundi2019}, TEM \cite{Chaudhry2019}, as well as reinforcement learning (RL) based ones like ER \cite{Rolnick2018}, SER \cite{Isele2018} and MER \cite{Riemer2018}.
In addition generators are used by pseudo-rehearsal approaches like \cite{Shin2017} and \cite{Kamra2017}.
\\
Regularization-based methods add additional terms (\enquote{regularizers}) to the loss function that pursue different purposes.
For example, SSL \cite{Aljundi2018} adds terms to increase the sparsity of representations through lateral inhibition.
Others methods such as IMM \cite{Lee2017} merge the parameters of different tasks based on their statistical moments.
EWC \cite{Kirkpatrick2016} penalizes changes to neurons that are important to the previous task, based on the Fisher information matrix (FIM).
Synaptic intelligence \cite{Zenke2017} is pursuing a similar goal.
LwM \cite{Dhar2018} suggests to use an attention-based loss based on Grad-CAM \cite{Selvaraju2016}.
Other approaches as LwF \cite{Li2016} rely on knowledge distillation mechanisms.
\\
Lastly, parameter isolation methods represent a more structure-oriented approach.
Subsets of neurons are assigned to individual sub-task and subsequently protected, or even dynamically added, if necessary.
Different strategies are possible, such as routing within PathNet \cite{Fernando2017}, pruning of parameters in PackNet \cite{Mallya2017}, masking parameters \cite{Mallya2018} or using attention masks with HAT \cite{Serra2018}.
PNNs \cite{Rusu2016} freeze parameters of a previous sub-task and add new instances for new tasks.
ExpertGate \cite{Aljundi2016} is a mixture-of-ensembles-based approach where separate classifiers rely on a shared feature extractor.
\\
As pointed out in \cite{Pfuelb2019}, a significant subset of these methods requires specific experimental setups to perform accurately, which may not be available in application settings.
For example, some methods require access to samples from future sub-tasks for tuning hyper-parameters, whereas others need to store all samples from past tasks for determining when to stop training.
Many proposed methods have a time and/or memory complexity that scales at least linearly with the number of tasks and thus may fail if this number is large.

\subsection{Goals and contributions}\label{sec:int:goa}
The following novel and relevant contributions are made:
\begin{itemize}
    \item an empirical investigation and comparison of recent replay-based CL approaches
    \item a proposal and systematic evaluation of elementary sample-selection strategies
    \item a comparison of evaluation metrics that enforce application constraints
    \item an in-depth discussion of advantages and disadvantages of different replay methods
\end{itemize}

\section{Methods}\label{sec:met}
As stated earlier, the two main branches in replay-based CL are represented by \textbf{rehearsal} and \textbf{constraint}-based methods.
Both of them make use of a \textbf{sample buffer} if no generators are involved.
We restrict ourselves to a more challenging setup with single-head output models, as well as class-incremental instead domain-incremental tasks.
Since multi-head outputs (e.g. task-incremental tasks in particular) would imply that the sub-task provenance of a sample must be known at evaluation/application time \cite{Ven2018, Hsu2018}.

\subsection{Investigated aspects}\label{sec:met:asp}
\begin{figure}[!h]
    \centering
    \includegraphics[width=\linewidth]{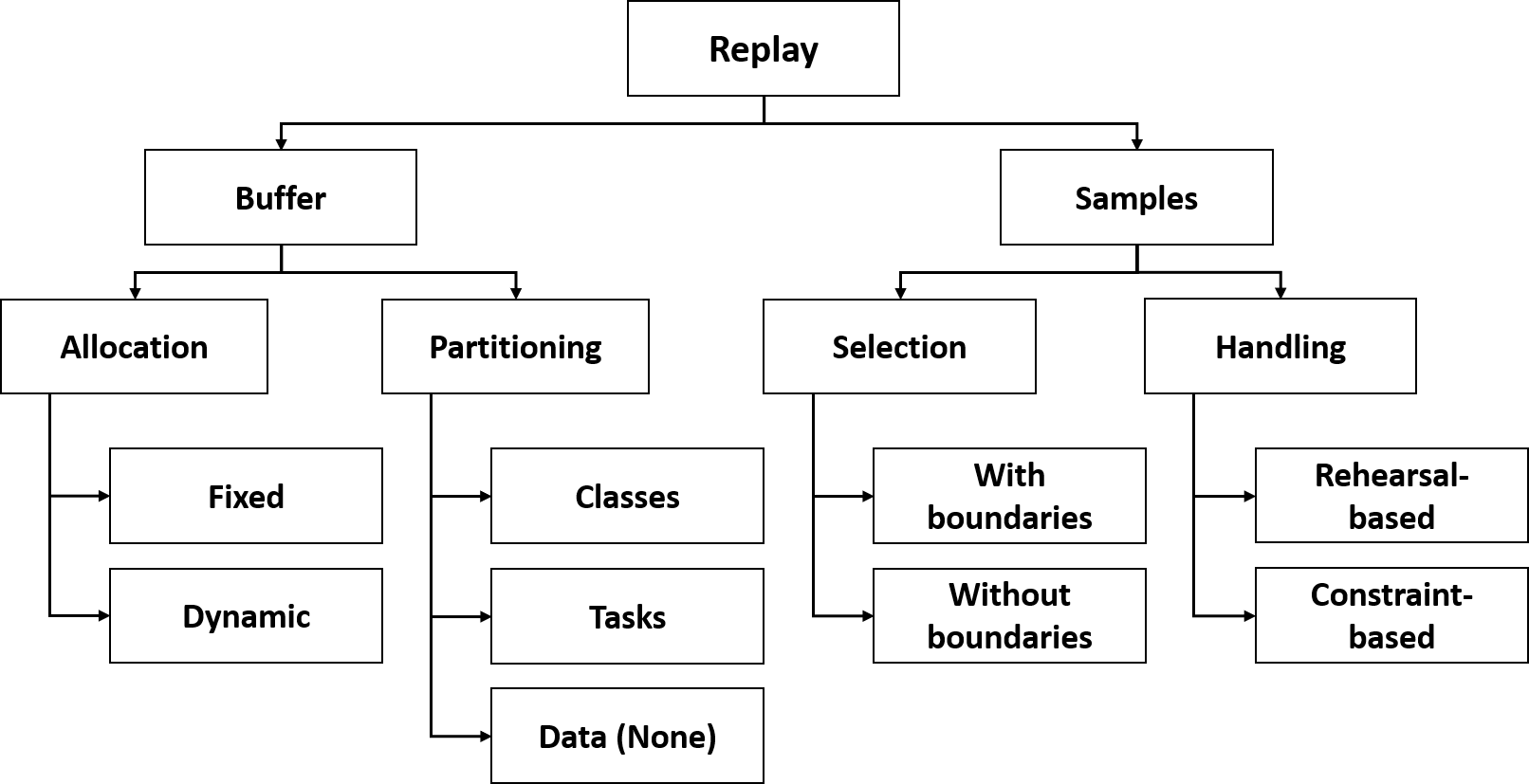}
    \caption{Overview of aspects that each replay-based approach must address.}
    \label{fig:overview_replay}
\end{figure}

Figure \ref{fig:overview_replay} contains all important aspects of replay-based approaches.
However, this article only investigates replay methods w.r.t. their sample strategies.
Two kinds of sample strategies can be distinguished regarding their meaning in relation to the buffer: the sample selection strategy as well as the sample handling strategy.
\par\noindent The \textbf{sample selection strategy} defines two criteria:
\begin{itemize}
    \item which sample is a potential candidate to be included in the buffer (referred-to as \textit{sample-in})
    \item which sample is a potential candidate to be removed from the buffer (referred-to as \textit{sample-out})
\end{itemize}
\par\noindent The \textbf{sample handling strategy} defines how the stored samples are employed for CL.
In particular:
\begin{itemize}
    \item which samples of the buffer could be used for replay
    \item how precisely the selected samples are used for replay
\end{itemize}
Due to a finite sized buffer, each later sample-in event requires a prior sample-out event.

\subsection{Investigated Models}\label{sec:met:mod}
\begin{table}[!h]
    \caption{The investigated approaches with their key characteristics.}
    \centering
    \resizebox{\linewidth}{!}{
        \begin{tabular}{c|c|c|c}
            \hline
            approach & replay-type & task-boundaries & task-types \\
            \hline\hline
            iCaRL \cite{Rebuffi2016} & rehearsal & yes & classes \\
            (A-)GEM \cite{LopezPaz2017, Chaudhry2018} & constraint & yes & classes \\
            TEM \cite{Chaudhry2019} & rehearsal & yes & classes \\
            GBSS \cite{Aljundi2019} & rehearsal & no & classes \\
            \hline
            \textbf{NSR} (ours) & rehearsal & yes & classes \\
            \textbf{NSR+} (ours) & rehearsal & no & classes \\
            \hline
        \end{tabular}
    }
    \label{tab:approach_overview}
\end{table}

\par\noindent\textbf{iCaRL} \cite{Rebuffi2016} is a rehearsal-based approach comprising a variety of concepts.
Sample selection is based on a herding algorithm, which assumes that all data samples of a sub-task are simultaneously available.
Concerning sample handling: the average feature representation of the penultimate layer is computed class-wise over the selected samples.
Finally these \enquote{means-of-exemplars} are used to calculate the predictions.
For each further sub-task, all existing class means must be re-calculated.
During the training process, the selected samples are also used for rehearsal purposes.
Furthermore, knowledge distillation is performed through the whole training process.
\par\noindent\textbf{(A-)GEM} \cite{LopezPaz2017, Chaudhry2018} is a constraint-based CL approach.
From each sub-task, the last $n$ samples are chosen, hence sample selection works as ring buffer.
Sample handling in GEM uses samples from the buffer to calculate their gradients and span a convex polyhedral cone in parameter space.
This cone defines a feasible region for loss gradients of the current sub-task, which are computed in a conventional DNN-like manner.
If the actual gradient is not in this subspace, a projection is computed by solving a quadratic optimization problem.
The sample handling of A-GEM avoids this complex computation and simply uses the mean gradient of such samples from the buffer.
If they point in the same direction the current gradient is applied, otherwise a orthogonal projection to the averaged gradient will be performed.
\par\noindent\textbf{TEM} \cite{Chaudhry2019} presents a comparison of different sample selection strategies for small-sized buffers.
The tested strategies are: reservoir sampling, ring-buffer (as (A-)GEM), k-means and Mean of Features (MoF) (like iCaRL).
Both reservoir sampling and the ring-buffer are efficient options to select samples in a random manner, whereas k-means and MoF are much more computationally intense selections.
Concerning sample handling: each mini-batch merges samples from the current sub-task with stored samples to obtain a averaged gradient.
Through the concept of mixed batch gradient updates, rehearsal-based approaches do not adversely affect the generalization capability of a model.
\par\noindent\textbf{GBSS} \cite{Aljundi2019} is a sample selection strategy for a setup without of task boundaries or at least knowledge about these.
Each seen sample is regarded as an individual constraint, to which every following sample must be compatible.
As in (A-)GEM, this means that their gradients should not point in the opposite direction.
Sample selection in this context is identical to a constraint reduction problem which is solved by a greedy strategy as surrogate for the actual quadratic optimization problem.
This strategy selects $n$ random samples from the buffer and calculates the cosine-similarity between the gradient of the current sample and the gradients of the selected samples.
Samples of the buffer are only replaced if the similarity falls below a defined threshold.
In this case, the sample with the maximal cosine-similarity is replaced.
Sample handling is comparable with TEM and realized by mixed batches.
\par\noindent With \textbf{Naive Sample Rehearsal (NSR)} we introduce a rehearsal-based approach and propose multiple sample selection strategies.
Based on the information of task boundaries, it is possible to implement several strategies: $n$ random, $n$ averaged, min/max intensity, nearest/farthest mean, lowest/highest variance.
\enquote{Random} selection chooses $n$ samples per class from each sub-task, while \enquote{averaged} selection splits the sub-task data into $n$ random subsets per class and calculates the averages over these.
\enquote{Intensity}-based sample selection determines the sum over all pixels and stores samples with the lowest or the highest value.
With \enquote{mean}-based selection, the difference of each sample to the corresponding class mean is calculated.
Samples are selected when they are either close to or far away.
Analogously, the squared differences are used for \enquote{variance}-based selection.
As for sample handling: the selected samples are stored in a buffer, which is merged with new sub-task data to train a neural network.
\par\noindent Through the expansion \textbf{Naive Sample Rehearsal Plus (NSR+)} we propose another rehearsal-based approach with different sample selection strategies.
Contrary to NSR strategies of NSR+ work in a more restrictive setup and without knowledge about task boundaries.
Implemented sample selection strategies are: random, false/true classification and min/max prediction.
The \enquote{random} strategy makes use of the available label information to randomly store samples per class and ensure that these are equally distributed.
The \enquote{false/true classification} strategy uses the predicted value of the true label to store samples: either the worst ones (for an incorrect classification) or the best ones (for a correct classification).
In contrast, the \enquote{min/max prediction} strategy only uses the logits of the classification layer, again retaining samples with best (strongest) and the worst (weakest) predictions, but based on a peak-to-peak measure.
For sample handling, the selected samples are combined with the new sub-task data batch-wise, because task boundaries are assumed to be unavailable.

\subsection{Datasets}\label{sec:exp:dat}
All seven approaches that we introduced in section \ref{sec:met:mod} are evaluated on four well-known benchmarks: MNIST \cite{Lecun1998}, Fashion-MNIST \cite{Xiao2017}, CIFAR10 \cite{Krizhevsky2009} and SVHN \cite{Netzer2011}.
They are frequently used in the research area of CL and contain $10$ classes in total.
Although these datasets are relatively simple w.r.t. non-continual classifier learning, this setup has been proven to be a very challenging one for CL, see \cite{Pfuelb2019}.
\\
MNIST and Fashion-MNIST both contain $60000$ grey-scaled images of shape $28 \times 28$ for the training.
MNIST represents handwritten digits from 0 to 9 while Fashion-MNIST consists of 10 different articles of clothing.
Unlike Fashion-MNIST, the MNIST dataset is not precisely balanced across classes.
\\
CIFAR10 and SVHN have RGB channels over the spatial domain of all images, hence the sample shapes of both are $32 \times 32 \times 3$.
CIFAR10 contains $50000$ training samples, of which 4 represent vehicles and 6 animals.
SVHN represents $73257$ digits from 0 to 9 taken from images of house numbers.
Similar to the grey-scale datasets, CIFAR10 is balanced across classes while SVHN is not
Samples of SVHN can have multiple digits of which only the centered one is relevant.

\subsection{Tasks}\label{sec:met:tas}
\begin{table}[!h]
    \caption{Description of class-incremental continual learning tasks with disjoints splits of classes into sub-tasks (ST).}
    \centering
    \resizebox{\linewidth}{!}{
        \begin{tabular}{l|ccccccc}
            \hline
            ID & ST 1 & ST 2 & ST 3 & ST 4 & ST 5 & \ldots & ST 10 \\
            \hline\hline
            $D1_{10}$ & 0-9 & - & - & - & - & \ldots & - \\
            $D2_{5}$ & 0-4 & 5-9 & - & - & - & \ldots & - \\
            $D5_{2}$ & 0,1 & 2,3 & 4,5 & 6,7 & 8,9 & \ldots & - \\
            $D10_{1}$ & 0 & 1 & 2 & 3 & 4 & \ldots & 9 \\
            \hline
        \end{tabular}
    }
    \label{tab:task_overview}
\end{table}

In this article, we analyse class-incremental continual learning tasks: sub-tasks consist of disjoint sets of classes from the vision benchmarks in section \ref{sec:exp:dat}.
See table \ref{tab:task_overview} for all used CL tasks.
Please note, that our sub-tasks contain an equivalent number of classes, but the number of samples per class can be slightly unbalanced, depending on the dataset.
\\
Disjoint sub-tasks are one of the most relevant application scenarios.
Assuming an arbitrary number of classes has already been trained, it should be subsequently possible to learn further classes, without the risk of performance degradation.
Additional such tasks are more difficult as domain-incremental ones, because the classes are completely new and not just another representation of already known ones \cite{Ven2018, Hsu2018}.

\subsection{Metrics}\label{sec:met:met}
Recent work mainly focuses on the final classification accuracy or the averaged sub-task accuracies.
Such measures are intuitive, but their usage shows limitations.
In particular, they do not express the mitigation of CF, or the effectiveness of transfer learning.
Hence, additional performance metrics may be required.
In this article, we rely on the forgetting measure proposed in \cite{Chaudhry2017} as well as on the forward and backward transfer measures from \cite{LopezPaz2017}.
Initially, a model is tested directly after the random initialization, before any sub-task is processed.
Subsequently, the model is tested after each following sub-task (or even, e.g., after each mini-batch within a sub-task).
Based on this, a matrix $R_{i, j}$ can be generated \cite{LopezPaz2017}, whose rows represent a time of measurement $i$ (e.g., mini-batch iteration), and whose columns represent the classification accuracies per measurement unit $j$ (e.g., per class).
This matrix has at least dimensions of $R \in \mathbb{R}^{T+1, T}$.
Where $T$ is the total number of tasks, if the evaluation is done after each sub-task and the initial evaluation is included.
Let be the indices $i \in \{0, \ldots, T\}$ and $j \in \{1, \ldots, T\}$.
\\
\\
The \textit{average accuracy} ($A \in [0, 1]$) for sub-task $k$, with $1 \leq k \leq T$ is defined as:
\begin{equation}
    A_{k} = \frac{1}{k} \sum_{j = 1}^{k} R_{k, j}
\end{equation}
$A_{T}$ represents the final accuracy which is the average accuracy over all sub-tasks after completion of training.
\\
For sub-task $k$, the \textit{forgetting measure} ($F \in [-1, 1]$), the \textit{forward transfer measure} ($FT \in [-1, 1]$) and the \textit{backward transfer measure} ($BT \in [-1, 1]$)  are defined as:
\begin{align}
    F_{k} &= \frac{1}{k - 1} \sum_{j = 1}^{k - 1} \max_{l \in \{1, \ldots, k - 1\}} R_{l, j} - R_{k, j}
    \\
    FT_{k} &= \frac{1}{T - k} \sum_{j = 1}^{T - k} R_{k, k + j} - R_{k - 1, k + j}
    \\
    BT_{k} &= \frac{1}{k - 1} \sum_{j = 1}^{k - 1} R_{k, j} - R_{k - 1, j}
\end{align}
Defining the forgetting measure on the initial sub-task is meaningless, just like evaluating $FT$ on the last sub-task and $BT$ on the first sub-task.

\begin{table}[!h]
    \caption{Visualization of possible {\color{red}$FT \downarrow$} and {\color{green}$BT \uparrow$} in evaluation matrix $R \in \mathbb{R}^{T+1,T}$, with $T=5$ as total number of sub-tasks.}
    \centering
    \resizebox{0.85\linewidth}{!}{
        \begin{tabular}{c|c c c c c}
            i / j & $j=1$ & $j=2$ & $j=3$ & $j=4$ & $j=5$ \\
            \hline
            $i=0$ & $R_{0,1}$ & $R_{0,2}$ & $R_{0,3}$ & $R_{0,4}$ & $R_{0,5}$ \\
            &  & $\color{red}\downarrow$ & $\color{red}\downarrow$ & $\color{red}\downarrow$ & $\color{red}\downarrow$ \\
            $i=1$ & $\mathbf{R_{1,1}}$ & $R_{1,2}$ & $R_{1,3}$ & $R_{1,4}$ & $R_{1,5}$ \\
            & $\color{green}\uparrow$ &  & $\color{red}\downarrow$ & $\color{red}\downarrow$ & $\color{red}\downarrow$ \\
            $i=2$ & $R_{2,1}$ & $\mathbf{R_{2,2}}$ & $R_{2,3}$ & $R_{2,4}$ & $R_{2,5}$ \\
            & $\color{green}\uparrow$ & $\color{green}\uparrow$ &  & $\color{red}\downarrow$ & $\color{red}\downarrow$ \\
            $i=3$ & $R_{3,1}$ & $R_{3,2}$ & $\mathbf{R_{3,3}}$ & $R_{3,4}$ & $R_{3,5}$ \\
            & $\color{green}\uparrow$ & $\color{green}\uparrow$ & $\color{green}\uparrow$ &  & $\color{red}\downarrow$ \\
            $i=4$ & $R_{4,1}$ & $R_{4,2}$ & $R_{4,3}$ & $\mathbf{R_{4,4}}$ & $R_{4,5}$ \\
            & $\color{green}\uparrow$ & $\color{green}\uparrow$ & $\color{green}\uparrow$ & $\color{green}\uparrow$ &  \\
            $i=5$ & $R_{5,1}$ & $R_{5,2}$ & $R_{5,3}$ & $R_{5,4}$ & $\mathbf{R_{5,5}}$ \\
        \end{tabular}
    }
    \label{tab:FT_and_BT}
\end{table}

Note that our versions of $FT$ and $BT$ are defined w.r.t. previously obtained values and not to the fixed initial accuracies ($\vec{b_{i}}$) or to the sub-task accuracies ($R_{i, i}$) itself, as in \cite{LopezPaz2017}.
Therefore, the number of sub-tasks for which $FT$ has a meaning is decreasing over time, while the number of sub-tasks for which $BT$ has a meaning is increasing.
Hence, $FT$ and $BT$ have to be computed for each sub-task to precisely catch accuracy changes caused by new sub-tasks.
As key figure, a weighted mean over all calculated $FT_{k}$ and $BT_{k}$ per experiment is computed.

\section{Experiments}\label{sec:exp}
A common property of replay-based approaches is that they use conventional DNN or convolutional neural network (CNN) architectures as \enquote{back-ends}.
The ability to continually learn in rehearsal approaches is mainly realized by controlling the samples that are fed into this back-end.

\subsection{Back-ends}\label{sec:exp:bac}
All back-ends should be executable with justifiable resources and practicable run-times.
Thus four different architectures are chosen to evaluate each approach.
Two of them are simple DNNs, one with a single hidden layer of $100$ neurons and another with $3$ hidden layers and $400$ neurons per hidden layer.
The others are CNN-based back-ends, represented by adapted versions of LeNet-5 \cite{Lecun1998} and VGG13 \cite{Simonyan2014}.
For all approaches, we use stochastic gradient descent (SGD) with momentum as optimizer, cross-entropy as loss function and softmax as normalization function (binary cross-entropy and sigmoid for iCaRL according to \cite{Rebuffi2016}).

\begin{table}[!h]
    \caption{List of evaluated hyper-parameters with their values.}
    \centering
    \resizebox{\linewidth}{!}{
        \begin{tabular}{l|l|l}
            \hline
            approaches & hyper-parameter & values \\
            \hline\hline
            & epochs & 1, 3, 5 \\
            & batch size & 10, 50, 250 \\
            & learning rate & 1e-2, 1e-3, 1e-4 \\
            & buffer size & 50, 200, 500 \\
            \hline
            TEM, GBSS, NSR+ & iterations & 1, 3, 5 \\
            GEM, GBSS & memory strength & 0.25, 0.5, 0.75 \\
            GBSS & threshold & -0.1, 0, +0.1 \\
            \hline
        \end{tabular}
    }
    \label{tab:hyper-parameters}
\end{table}

Table \ref{tab:hyper-parameters} presents the considered hyper-parameters of our investigation.
Some of them are approach independent, others not.
All missing approach specific hyper-parameters are used with their recommended values and ignored for the most part.
Not included are also (hyper-)parameters regarding the protocols (seed, model, network, dataset and task), as well as default parameters (e.g. selection strategy).

\subsection{Baselines}\label{sec:exp:bas}
As indicated in section \ref{sec:met:met} it is already challenging to measure the performance of replay approaches.
Additionally, the obtained measurements should be compared with meaningful baselines.
We use three different baselines for the evaluation presented in this section: \textit{Joint Training} (JT), \textit{Raw-Task} (RT) and \textit{Raw-Buffer} (RB).

\begin{table*}[!hb]
    \caption{All baselines: joint training (JT), raw-task (RT) and raw-buffer (RB) w.r.t. our datasets, tasks and buffers.}
    \centering
    \resizebox{\textwidth}{!}{
        \begin{tabular}{l|l|c|c|c|c|c|c|c|c|c|c|c|c}
            \hline
            && \multicolumn{3}{c|}{MNIST} & \multicolumn{3}{c|}{Fashion-MNIST} & \multicolumn{3}{c|}{CIFAR10} & \multicolumn{3}{c}{SVHN} \\
            task & buffer & JT & RT & RB & JT & RT & RB & JT & RT & RB & JT & RT & RB \\
            \hline
            \hline
            $D2_{5}$ & iCaRL & \multirow{21}{*}{$0.9832$} & \multirow{7}{*}{$0.4998$} & $0.5812$ & \multirow{21}{*}{$0.8653$} & \multirow{7}{*}{$0.5293$} & $0.6180$ & \multirow{21}{*}{$0.4907$} & \multirow{7}{*}{$0.3244$} & $0.2083$ & \multirow{21}{*}{$0.7240$} & \multirow{7}{*}{$0.4327$} & $0.1201$ \\
            & GEM &&& $0.2819$ &&& $0.5486$ &&& $0.2146$ &&& $0.1045$ \\
            & A-GEM &&& $0.3055$ &&& $0.5843$ &&& $0.2570$ &&& $0.1184$ \\
            & TEM &&& $0.3264$ &&& $0.5052$ &&& $0.2156$ &&& $0.1148$ \\
            & GBSS &&& $0.2954$ &&& $0.4098$ &&& $\textit{0.1000}$ &&& $\textit{0.1000}$ \\
            & NSR &&& $0.6025$ &&& $0.6518$ &&& $0.2408$ &&& $0.1340$ \\
            & NSR+ &&& $0.4513$ &&& $0.6022$ &&& $0.2281$ &&& $0.1038$ \\
            $D5_{2}$ & iCaRL && \multirow{7}{*}{$0.1987$} & $0.5212$ && \multirow{7}{*}{$0.1994$} & $0.6236$ && \multirow{7}{*}{$\textit{0.1000}$} & $0.2148$ && \multirow{7}{*}{$\textit{0.1000}$} & $0.1142$ \\
            & GEM &&& $0.3293$ &&& $0.5120$ &&& $0.2342$ &&& $0.1004$ \\
            & A-GEM &&& $0.3131$ &&& $0.5216$ &&& $0.2261$ &&& $0.1081$ \\
            & TEM &&& $0.3554$ &&& $0.5349$ &&& $0.2301$ &&& $0.1065$ \\
            & GBSS &&& $0.1804$ &&& $0.2462$ &&& $\textit{0.1000}$ &&& $\textit{0.1000}$ \\
            & NSR &&& $0.5742$ &&& $0.6430$ &&& $0.2372$ &&& $0.1238$ \\
            & NSR+ &&& $0.3515$ &&& $0.5219$ &&& $0.2109$ &&& $0.1070$ \\
            $D10_{1}$ & iCaRL && \multirow{7}{*}{$0.1000$} & $0.5602$ && \multirow{7}{*}{$0.1000$} & $0.6405$ && \multirow{7}{*}{$0.1000$} & $0.2254$ && \multirow{7}{*}{$0.1000$} & $0.1192$ \\
            & GEM &&& $0.3541$ &&& $0.5462$ &&& $0.2315$ &&& $0.1071$ \\
            & A-GEM &&& $0.3842$ &&& $0.5843$ &&& $0.2190$ &&& $0.1076$ \\
            & TEM &&& $0.3562$ &&& $0.5248$ &&& $0.2272$ &&& $0.1225$ \\
            & GBSS &&& $\textit{0.1000}$ &&& $\textit{0.1000}$ &&& $\textit{0.1000}$ &&& $\textit{0.1000}$ \\
            & NSR &&& $0.5975$ &&& $0.6673$ &&& $0.2398$ &&& $0.1302$ \\
            & NSR+ &&& $0.2756$ &&& $0.4598$ &&& $0.2214$ &&& $\textit{0.1000}$ \\
            \hline
        \end{tabular}
    }
    \label{tab:baselines}
\end{table*}

JT is the most common baseline for CL and often seen as an upper bound for any CL approach, indicating the maximal performance that can be achieved.
This seems reasonable, although there is no formal proof, and situations might exist where the sequential presentation of sub-tasks enhances performance.
In CL settings, RT is the naive baseline performance of the \enquote{raw} model trained sequentially on all sub-tasks without any additional mechanisms like, e.g., fine-tuning or knowledge distillation.
It can thus be considered as a (probably very loose) lower bound.
Similarly, we introduce RB as another baseline of the \enquote{raw} model, but in particular for replay.
Differently from RT, this model is jointly trained for each approach on all selected samples of their buffers and tested on the whole dataset.
The RB measurement allows a better evaluation of all different sample selection strategies and is inspired by the \textit{Fitting Capacity} measure for generator-based replay approaches \cite{Lesort2018}.
These three baselines significantly facilitate the comparability of replay-based approaches.

\subsection{Additional experiments}\label{sec:exp:add}
Beside the protocols described in section \ref{sec:met:tas}, additional experiments are performed.
In essence, we focus on three aspects and examine their interrelationship:
\par\noindent\textbf{Online/offline} First we want to analyse how the approaches perform in an online setup, where each sample is seen once.
In this case, only a single pass over the sub-task data or a single iteration over a mini-batch is permitted.
In contrast, an offline setup is used to analyze whether the approaches can reach higher performance if they are able to process samples multiple times.
Both setups are independent of whether task boundaries are given or not.
\par\noindent\textbf{Rehearsal/no rehearsal} Both GEM and A-GEM are fully constraint-based approaches, which is why they can be investigated with an additional repetition of their buffers.
Currently, samples are used to generate constraints w.r.t. gradients, but without any type of sample.
Due to the fact, that they are already existing and used during the training process, it is interesting to evaluate whether the rehearsal affects the learning process of constraint-based methods.
\par\noindent\textbf{Over-sampling/non over-sampling} The kind of rehearsal differs between evaluated methods.
Certain approaches (TEM, GBSS, NSR+) merge samples of the current sub-task in an 1:1 ratio with samples from the buffer per mini-batch.
Other ones (iCaRL, NSR) usually merge all samples from the buffer with the current sub-task data only once (without over-sampling).
However, it is also possible to over-sample from their buffers and we will compare both options.
\\
The impact of all additional evaluation protocols, as well as the outcomes, will be discussed in section \ref{sec:dis}.

\subsection{Results}\label{sec:exp:res}
\begin{table*}[!ht]
    \caption{Final accuracy/forgetting measures; averages over 5 runs for the optimal hyper-parameter setup of each approach.}
    \centering
    \resizebox{\textwidth}{!}{
        \begin{tabular}{l|l|c|c|c|c|c|c|c|c|c|c|c|c|c|c}
            \hline
            && \multicolumn{2}{c|}{iCaRL} & \multicolumn{2}{c|}{GEM} & \multicolumn{2}{c|}{A-GEM} & \multicolumn{2}{c|}{TEM} & \multicolumn{2}{c|}{GBSS} & \multicolumn{2}{c|}{NSR} & \multicolumn{2}{c}{NSR+} \\
            sub-task & benchmark & $A_{T}$ & $F_{T}$ & $A_{T}$ & $F_{T}$ & $A_{T}$ & $F_{T}$ & $A_{T}$ & $F_{T}$ & $A_{T}$ & $F_{T}$ & $A_{T}$ & $F_{T}$ & $A_{T}$ & $F_{T}$ \\
            \hline\hline
            $D2_{5}$
            & MNIST & $0.7964$ & $+0.0849$ & $0.9313$ & $+0.1115$ & $0.9254$ & $+0.1039$ & $0.9074$ & $+0.0713$ & $0.8422$ & $+0.1205$ & $0.9406$ & $+0.0939$ & $\textbf{0.9633}$ & $\textbf{+0.0510}$ \\
            & F.-MNIST & $0.6422$ & $\textbf{+0.0358}$ & $0.8022$ & $+0.2972$ & $0.7944$ & $+0.0490$ & $0.7977$ & $+0.2941$ & $0.8027$ & $+0.1303$ & $0.8030$ & $+0.0572$ & $\textbf{0.8318}$ & $+0.1122$ \\
            & CIFAR10 & $0.2810$ & $\textbf{+0.1926}$ & $0.3531$ & $+0.5768$ & $0.3584$ & $+0.5943$ & $\textbf{0.3950}$ & $+0.5615$ & $0.2593$ & $+0.5136$ & $0.3839$ & $+0.6120$ & $0.3601$ & $+0.5977$ \\
            & SVHN & $0.2033$ & $\textbf{+0.1023}$ & $0.6007$ & $+0.4396$ & $0.5641$ & $+0.4215$ & $0.6743$ & $+0.3555$ & $0.5950$ & $+0.1866$ & $0.5832$ & $+0.4692$ & $\textbf{0.7060}$ & $+0.2178$ \\
            $D5_{2}$
            & MNIST & $0.6687$ & $+0.1615$ & $\textbf{0.9327}$ & $+0.0925$ & $0.8990$ & $+0.1218$ & $0.8640$ & $+0.0846$ & $0.6119$ & $+0.2902$ & $0.9274$ & $\textbf{+0.0843}$ & $0.9308$ & $+0.0869$ \\
            & F.-MNIST & $0.6823$ & $+0.1485$ & $\textbf{0.7757}$ & $\textbf{+0.0662}$ & $0.6173$ & $+0.3293$ & $0.7410$ & $+0.3154$ & $0.5024$ & $+0.3451$ & $0.7593$ & $+0.0919$ & $0.7333$ & $+0.1971$ \\
            & CIFAR10 & $0.2759$ & $\textbf{+0.1563}$ & $0.2993$ & $+0.4374$ & $0.2805$ & $+0.4708$ & $\textbf{0.3107}$ & $+0.4896$ & $0.2104$ & $+0.4054$ & $0.2871$ & $+0.4696$ & $0.2979$ & $+0.4502$ \\
            & SVHN & $0.1951$ & $+0.2034$ & $\textit{0.1000}$ & $\textit{0.0000}$ & $\textit{0.1000}$ & $\textit{0.0000}$ & $\textit{0.1000}$ & $\textit{0.0000}$ & $0.1928$ & $+0.1468$ & $\textbf{0.3336}$ & $\textbf{-0.0530}$ & $\textit{0.1000}$ & $\textit{0.0000}$ \\
            $D10_{1}$
            & MNIST & $0.7449$ & $+0.1083$ & $0.9133$ & $+0.0763$ & $0.8512$ & $+0.0807$ & $0.8611$ & $\textbf{+0.0507}$ & $0.6642$ & $+0.2074$ & $\textbf{0.9197}$ & $+0.0718$ & $0.8620$ & $+0.0581$ \\
            & F.-MNIST & $0.6536$ & $+0.1287$ & $0.7753$ & $\textbf{-0.0290}$ & $0.6581$ & $+0.1297$ & $\textbf{0.7776}$ & $+0.1989$ & $0.3767$ & $+0.4380$ & $0.7750$ & $+0.0496$ & $0.7601$ & $+0.1394$ \\
            & CIFAR10 & $0.2696$ & $+0.2898$ & $\textbf{0.2994}$ & $+0.2711$ & $0.2133$ & $+0.2430$ & $\textit{0.1000}$ & $\textit{0.0000}$ & $\textit{0.1000}$ & $\textit{0.0000}$ & $0.2909$ & $\textbf{+0.2130}$ & $\textit{0.1000}$ & $\textit{0.0000}$ \\
            & SVHN & $0.2017$ & $+0.1737$ & $\textbf{0.3579}$ & $+0.2250$ & $0.2649$ & $+0.3187$ & $\textit{0.1000}$ & $\textit{0.0000}$ & $\textit{0.1000}$ & $\textit{0.0000}$ & $0.3460$ & $\textbf{-0.1772}$ & $\textit{0.1000}$ & $\textit{0.0000}$ \\
            \hline
        \end{tabular}
    }
    \label{tab:best_results}
\end{table*}

Table \ref{tab:best_results} shows the overall best results determined by the maximal final accuracy denoted as $A_{T}$.
It is not surprising that all of them are based on the largest evaluated buffer size.
All investigated replay approaches perform can mitigate CF.
Depending on the specific sub-task, some accuracy measures ($A_{T}$) are nearly as good as the JT baseline (compare table \ref{tab:baselines}).
In general, the performance is decreasing with the number of sub-tasks as well as the difficulty of learning problems.
Some experiments are not able to reach significant accuracies at all.
The final forgetting measure $F_{T}$ provides more detailed insights: CIFAR10 and SVHN generally show stronger forgetting ($F_{T}$) and the forgetting of tasks containing less classes is better.
Only iCaRL behaves differently in this respect.

\begin{table*}[!hb]
    \caption{The best $FT$s and $BT$s of all evaluated experiments for the three tasks.}
    \centering
    \resizebox{\textwidth}{!}{
        \begin{tabular}{l|c|c|c|c|c|c|c|c|c|c|c|c|c|c}
            \hline
            & \multicolumn{2}{c|}{iCaRL} & \multicolumn{2}{c|}{GEM} & \multicolumn{2}{c|}{A-GEM} & \multicolumn{2}{c|}{TEM} & \multicolumn{2}{c|}{GBSS} & \multicolumn{2}{c|}{NSR} & \multicolumn{2}{c}{NSR+} \\
            task & $FT$ & $BT$ & $FT$ & $BT$ & $FT$ & $BT$ & $FT$ & $BT$ & $FT$ & $BT$ & $FT$ & $BT$ & $FT$ & $BT$ \\
            \hline
            \hline
            $D2_{5}$
            & $\textbf{0.0000}$ & $-0.0322$ & $-0.0192$ & $+0.0838$ & $-0.0192$ & $0.0000$ & $-0.0192$ & $+0.0740$ & $-0.0192$ & $+0.0542$ & $-0.0192$ & $\textbf{+0.1187}$ & $-0.0192$ & $+0.1015$ \\
            & \textbf{MNIST} & F.-MNIST & F.-MNIST & SVHN & F.-MNIST & SVHN & F.-MNIST & F.-MNIST & F.-MNIST & F.-MNIST & F.-MNIST & \textbf{SVHN} & F.-MNIST & SVHN \\
            $D5_{2}$
            & $\textbf{0.0000}$ & $-0.0265$ & $-0.0065$ & $+0.1298$ & $-0.0065$ & $+0.0047$ & $-0.0065$ & $+0.0651$ & $-0.0065$ & $+0.0211$ & $-0.0065$ & $\textbf{+0.2333}$ & $-0.0065$ & $+0.0126$ \\
            & \textbf{MNIST} & SVHN & F.-MNIST & MNIST & F.-MNIST & F.-MNIST & F.-MNIST & F.-MNIST & F.-MNIST & F.-MNIST & F.-MNIST & \textbf{MNIST} & F.-MNIST & F.-MNIST \\
            $D10_{1}$
            & $\textbf{0.0000}$ & $-0.0164$ & $-0.0112$ & $\textbf{+0.1091}$ & $-0.0112$ & $+0.0244$ & $-0.0112$ & $+0.0916$ & $-0.0112$ & $+0.0495$ & $-0.0112$ & $+0.1040$ & $-0.0112$ & $+0.0561$ \\
            & \textbf{MNIST} & CIFAR10 & F.-MNIST & \textbf{MNIST} & F.-MNIST & MNIST & F.-MNIST & MNIST & F.-MNIST & MNIST & F.-MNIST & MNIST & F.-MNIST & F.-MNIST \\
            \hline
        \end{tabular}
    }
    \label{tab:best_tranfer}
\end{table*}

Due to the sequential training of sub-tasks, CL approaches may be able to achieve transfer between sub-tasks.
Transfer occurs if sub-tasks interfere with each other, which means they must share parameters.
The interferences are positive if both sub-tasks enforce similar parameter changes w.r.t. gradient updates.
Otherwise, the transfer is negative and causes a deterioration of old sub-task performances.
Since transfer is possible in both directions of the learning process ($k \prec t$ or $k \succ t$), $FT$ and $BT$ are required to describe it accurately (compare table \ref{tab:FT_and_BT}).
Precisely, $FT$ describes how the training of the current sub-task affects accuracies of further sub-tasks, whereas $BT$ describes these changes regarding all previous sub-tasks.
The best transfer measures for each approach are presented in table \ref{tab:best_tranfer}.
Clearly, disjoint class-incremental tasks cannot reach performances as high as domain-incremental tasks, which learn multiple transformations of the same objective.

\section{Discussion}\label{sec:dis}
The summarized results presented in the previous section are only a small subset of the experiments investigated, thus further outcomes and findings can be discussed.

\subsection{Outcomes}\label{sec:dis:out}
Unsurprisingly, the baselines confirm that some benchmarks (MNIST, Fashion-MNIST) are easier than others.
However, besides the overall difficulty of a benchmark, additional factors seem to be relevant for CL.
Comparing of different approaches (see table \ref{tab:best_results}), we can state the following:
\begin{itemize}
    \item iCaRL has stable forgetting measures, but cannot reach as high accuracies as other methods
    \item GEM can reach high final accuracies, as well as low forgetting measures over some experiments
    \item A-GEM performs slightly worse than GEM (but is of course much simpler to train)
    \item TEM reaches better results as A-GEM, hence the gradient update by mixed mini-batches is beneficial
    \item GBSS is based on GEM concepts, but in a more difficult setting, which results in a worse performance
    \item our proposals NSR(+) reach overall strong performances with appropriate sample selection strategies, as well as minimal computational cost
\end{itemize}
In general, approaches with explicit knowledge of task boundaries perform more consistently, and rehearsal-based approaches that over-sample from their buffer suffer more from unbalanced sub-tasks than others.
Regarding the transfer measures (compare table \ref{tab:best_tranfer}), we can observe further tendencies:
\begin{itemize}
    \item iCaRL has by design no $FT$ and only negative $BT$
    \item GEM reaches the highest $BT$ value for a $D10_{1}$ task and has many times positive $BT$
    \item A-GEM has nearly no $BT$, the measures are close to zero and not always positive
    \item TEM can achieve much better $BT$ measures as A-GEM, but is still worse than GEM
    \item GBSS $BT$ values are between A-GEM (as lower bound) and TEM (as upper bound)
    \item our approaches NSR(+) reach the best $BT$ for a $D2_{5}$ and a $D5_{2}$ task
\end{itemize}
All methods except iCaRL have almost identical negative $FT$ regarding the first sub-task.
During the whole training process, no change of $FT$ is observed (neither positive nor negative).
Some more or less unrelated but relevant findings are:
\begin{itemize}
    \item A higher number of epochs or iterations correlates with a better CL performance up to a certain point.
    Offline setups dominate online ones.
    \item (A-)GEM generally performs better with an additional repetition of stored samples.
    This applies for a task-wise, as well as a batch-wise rehearsal setting.
    \item The over-sampling of buffer contents has a positive effect.
    However, too high rehearsal rates can result in lower generalization and overfitting.
\end{itemize}
Depending on the sample handling strategy, the buffer itself can be divided into different partitions.
Possible and tested options are: none, by tasks or by classes.
For supervised learning scenarios, a by-class partition is a good choice since this is suitable for the most experiments.
Additionally, by construction it can be ensured that the samples are balanced over all classes as well.
At least the fixed size buffer should be filled dynamically, as already introduced by \cite{Rebuffi2016}.
This avoids the need for a priori knowledge about the number of sub-tasks.
\\
Regarding the buffer size, it is obvious that with smaller buffer sizes, the samples have to be selected with greater care to be as representative as possible.
Another observation is that samples with higher intensities (pixels different to background), higher deviations (compared to the class mean) and samples with the worst prediction (most uncertain distribution over logits) are also very rewarding w.r.t. the mitigation of forgetting.
\\
However, some investigated approaches use only a subset of the buffer instead of all selected samples.
In such case a random selection is performed to determine which ones are used for constraint or rehearsal purposes.

\subsection{Key Findings}\label{sec:dis:fin}
\par\noindent\textbf{Constraint-based approaches}
GEM requires a costly solution to a quadratic program, while A-GEM performs an orthogonal projection if the gradient violates the required constraint \cite{Chaudhry2018}.
From multiple experiments, it is possible to state that gradients violating this constraint do not interfere strongly with previous sub-tasks.
From the point of forgetting, this is desirable, since there is no performance loss w.r.t. previous sub-tasks, but positive transfer is limited as well.
Additionally, the projection may prevent proper learning of the current sub-task.
A-GEM's projection is not only highly restrictive but computationally costly compared to NSR or NSR+.
\par\noindent\textbf{Rehearsal-based approaches}
An alternative to A-GEM is given by merging stored samples with samples of the current sub-task.
Both approaches are similar, but rehearsal is frequently the better solution, as the results show.
Contrary to A-GEM, the resulting gradient is automatically calculated by SGD updates and can only be manipulated implicitly over the ratio between old and new sub-task samples per batch.
This is because the mean gradient is a trade-off between the direction of old sub-tasks samples and samples of the new sub-task as shown by \cite{Chaudhry2019}.
While A-GEM uses the plain gradient, if it is compatible and only projects, if it violates the constraint, these methods always perform identical.
This may affect the convergence of sub-tasks due to the continuous and strong binding to samples of previous sub-tasks.
\par\noindent\textbf{Task boundaries}
CL without knowledge about task boundaries is much more challenging.
However, the main issue is the balanced selection of representative samples over all classes seen so far.
Considering the values of our RB baseline in table \ref{tab:baselines}, the difference between GBSS and NSR+ is simply the sample selection strategy and not the sample handling strategy.
If the distribution of the buffer cannot represent the data distribution, the method will miss its only purpose (to mitigate CF).
In our investigation, NSR+ outperforms GBSS, because the buffer captures the actual distribution more accurate.
\par\noindent\textbf{Replay issues}
Replay-based approaches construct combined optima (previous sub-tasks and current sub-task) or feasible regions by \enquote{valid} gradient directions, but their reachability is not ensured within a single sub-task.
This could be a major issue, unless there is one mitigating factor: during the further training process samples of previous sub-tasks are still replayed, which increases the accuracy of these sub-tasks afterwards.
However GEM, A-GEM, TEM and NSR+ fail for the SVHN benchmark at the $D5_{2}$ task as well as TEM, GBSS and NSR+ for CIFAR10 and SVHN at the $D10_{1}$ task.
Additionally, the learning process itself is unbalanced for the most replay-based approaches, since samples from earlier tasks are replayed more often then samples of later tasks.
Depending on individual approaches, this could create problems and has to be avoided by an additional weighting.

\subsection{Comparability and reproducibility}\label{sec:dis:com}
Past works compared multiple approaches, but ignored their conceptual intention, as already described by \cite{Prabhu2020} or \cite{Buzzega2020}.
In general it is difficult to draw comparisons to other works, because of the broad variability of CL settings.
iCaRL \cite{Rebuffi2016}, GEM \cite{LopezPaz2017}, A-GEM \cite{Chaudhry2018} and TEM \cite{Chaudhry2019} do not evaluate tasks comparable to ours, but the general tendencies are congruent with ours.
GBSS \cite{Aljundi2019} performs similar benchmarks as well as tasks, but their back-ends are different.
Additionally they train on a specific task size and consider only balanced tasks as well as balanced classes (per task).
This has a huge impact on such an approach, as we described in the previous section.
At least the works of \cite{Aljundi2019a, Buzzega2020, Jin2020, Prabhu2020} provide benchmarks on disjoint MNIST (split MNIST) and disjoint CIFAR10 (split CIFAR10).
Their findings are mostly consistent with ours.

\subsection{Current superiority}\label{sec:dis:sup}
The authors of \cite{Knoblauch2020} considered the complexity of CL for the first time from a theoretical point of view.
They demonstrated that CL is mappable to the satisfiability problem (SAT) and is therefore NP-hard.
In such a view, replay-based approaches come closest to their definition of optimal continual learning.
This is based on the \enquote{perfect memory} criterion, which means that CL algorithms can reconstruct at least one element (sample) from each equivalence set (sub-task).
\\
Through replay, it is possible to mitigate CF in various CL settings with a minimal effort.
Stored samples of a buffer can be used either for defining constraints or by rehearsal.
As shown in section \ref{sec:dis:fin}, these two variants have different effects.
However, both ensure a valid gradient update during each training step to optimize previous and current sub-tasks simultaneously; albeit this slows down the speed of convergence.
In addition, the concept of averaged gradients based on mixed mini-batches does not need a dedicated computation.
Contrary, approaches based on parameter isolation or regularization must perform difficult and expensive computations.
These are often not practical in real-world applications with certain constraints \cite{Pfuelb2019}.
In particular, processing data in real-time needs to be as fast as possible.

\subsection{Advantages and disadvantages}\label{sec:dis:adv}
Currently replay-based methods outperform other methods under a broad variety of application-oriented constraints.
Neither parameter isolation nor regularization-based approaches reach comparable performances.
Apart from that, the memory and computation overhead is small and of fixed complexity.
Replay approaches scale, in part, consistently across an arbitrary number of tasks.
Even embedded systems can use surplus memory as a buffer or generator to perform replay.
Another advantage is universality, which is why replay-based methods are quite generic and can be used in almost any scenario without conceptual changes.
Thus, it is possible to make existent models resistant to CF by adding replay abilities in the individual back-end.
Furthermore, methods are simple to implement and have a fundamental explainable conception.
The approach to use old sub-tasks samples during new sub-tasks to prevent forgetting is obvious.
They enforce some kind of combined optima, which is comparable to a joint training over sub-task seen so far (however such optima depend on subsets as well as stateful network parameters).
\\
On the other hand, there are immediate drawbacks that must be taken into account.
One unavoidable issue is the need of additional memory for the buffer, the size of which is an immediate trade-off between a better performance and a smaller memory footprint (compare section \ref{sec:dis:out} for details).
Contrary to other approaches, replay suffers from unbalanced learning processes.
This applies to both: intra as well as inter-task imbalances, where rehearsal-based methods are more affected than constraint-based ones.
Supposedly simple tasks, like learning all classes except one, which should be learned in a separate task, are difficult to handle.
By storing samples directly instead of using generative models, replay approaches may violate aspects of privacy, which is another disadvantage.
Depending on the processed data, this could be an intolerable fact especially if critical, e.g., medical or personal data is stored.
Lastly replay-based methods run into the risk of overfitting, if they are based on samples, which are not able to describe the data accurately.
Subsets have to be chosen carefully to adapt and represent the distribution of the data in the best possible way.
A too biased sample selection may affect the generalization ability of old sub-tasks equally, as well as the new sub-task (see section \ref{sec:dis:fin}).

\section{Conclusion and Outlook}\label{sec:con}
%

\subsection{Conclusion}\label{sec:con:con}
We introduce NSR(+), two rehearsal-based replay methods, which perform with less complexity at least as good as current state-of-the-art approaches.
Moreover, our work demonstrates that sample selection strategy is a key factor, in particular for small-sized buffers.
Of course, larger numbers and higher diversities of stored samples result in better performances.
The question of whether replay approaches should store samples directly or hold a generator that can produce any number of plausible samples is therefore not immediately clear.
Missing task boundaries require more sophisticated strategies due to the lack of information, but naive ones have also been introduced.
Additionally we propose a new baseline Raw-Buffer (RB) to measure and compare sample selection strategies.
Currently, the majority of investigated approaches are able to reliable generate balanced buffers for replay, with only a single approach failing to do this.
However, some of them suffer severely from unbalanced sub-task data, which is exacerbated by over-sampling.
Furthermore, we introduce modified metrics to measure transfer between single sub-tasks.
Transfer seems to be possible within our experiments, but not as much as for other learning problems.
None of the evaluated methods achieve $FT$ on our class-disjoint tasks.
Only the initial performance drop regarding the first sub-task is noticeable.
In contrast, $BT$ is present and also important for replay, because performances of earlier sub-tasks can still increase during the training process.
Hence, interpretations of $F$ must always be performed in combination with $BT$.

\subsection{Outlook}\label{sec:con:out}
Recent works introduced approaches which back-propagate the error of the loss function down to the input layer and modify the stored samples:
\cite{Aljundi2019a, Jin2020} increase the forgetting of stored samples, so that their rehearsal is more effective.
This approach attempts to prevent a model from forgetting sub-tasks by increasing the forgetting of individual samples in a particular way.
Other approaches \cite{Liu2020, Chaudhry2020} modify the stored samples to the effect that the boundaries between single sub-tasks are more explicit.
Sharp task separations have also a positive effect to CL and can mitigate forgetting.
\\
Further works \cite{Prabhu2020, Buzzega2020} have the objective of general continual learning.
They attempt to consider as few assumptions as possible, leading to very generic problem formulations.
General continual learning illustrates how broad the subject of CL has become nowadays.
Yet other works as \cite{Knoblauch2020} consider the theoretical requirements of optimal continual learning and point out the basic complexity of CL.
If an approach fulfills the perfect memory criterion, it has a fundamental prerequisite to perform well.
\\
Unbalanced data as well as unbalanced tasks are another topic to handle \cite{Kim2020}, because in these cases random or reservoir based methods will fail to effectively fill buffers.
Based on the selected samples, the buffer would adapt such an unbalanced distribution, which affects the learning behavior in addition.

{\small
    \bibliographystyle{IEEEtran}
    \bibliography{RBCL}
}

\end{document}